\documentclass[conference]{IEEEtran}
\IEEEoverridecommandlockouts

\DeclareRobustCommand\onedot{\futurelet\@let@token\@onedot}
\def\@onedot{\ifx\@let@token.\else.\null\fi\xspace}

\newcommand{\tableline}{\noalign{\hrule height 1.5pt}} 

\usepackage{array}

\newcolumntype{C}[1]{>{\centering\arraybackslash}m{#1}} %
\newcolumntype{L}[1]{>{\raggedright\arraybackslash}m{#1}}

\usepackage{cite}
\usepackage{comment}
\usepackage{amsmath}
\usepackage{amssymb}
\usepackage{graphicx}
\usepackage{booktabs}
\usepackage{graphicx}
\usepackage{multirow}

\DeclareMathOperator*{\argmin}{arg\,min}

\begin{document}

\title{Semantic-weighted ICP for LiDAR Odometry: Class-Aware Residual Reweighting for Robust Scan Registration}

\author{\IEEEauthorblockN{1\textsuperscript{st} Vasco Carvalho} 
\IEEEauthorblockA{\textit{Institute of Systems and Robotics}
\\
\IEEEauthorblockA{\textit{University of Coimbra}\\
Coimbra, Portugal\\
vasco.carvalho@isr.uc.pt}
}
\and
\IEEEauthorblockN{2\textsuperscript{nd} Tiago Barros} 
\IEEEauthorblockA{\textit{Institute of Systems and Robotics}
\\
\IEEEauthorblockA{\textit{University of Coimbra}\\
Coimbra, Portugal\\
tiagobarros@isr.uc.pt}
}
\and
\IEEEauthorblockN{3\textsuperscript{rd} Urbano J. Nunes} 
\IEEEauthorblockA{\textit{Institute of Systems and Robotics}
\\
\IEEEauthorblockA{\textit{University of Coimbra}\\
Coimbra, Portugal\\
urbano@isr.uc.pt}
}
}

\maketitle

\begin{abstract}

LiDAR odometry is a fundamental component of autonomous robotic systems, relying on geometric registration between consecutive point clouds to estimate ego-motion. However, traditional geometric approaches often degrade in dynamic or unstructured environments due to unreliable correspondences caused by moving objects, sparse geometric features, vegetation, and semantically ambiguous structures.  Existing works have shown that, some of these limitations can be addressed by introducing semantic information from the environment in the  registration process. 

In this work, we build on this, and show that not all elements in the environment are equally relevant for registration. Hence, we propose a semantic class-weighted ICP for LiDAR odometry. Instead of strictly filtering out points belonging to specific semantic classes, the proposed approach weights the residuals of points belonging to semantic categories based on their expected geometric stability. This strategy enables informative but potentially unstable structures, to contribute to the registration process while mitigating the influence of dynamic objects.

The experimental evaluation was conducted on the SemanticKITTI~\cite{behley2019iccv} and RELLIS-3D datasets, which include urban, highway, rural, and off-road environments. The empirical results show that the proposed Semantic-weighted ICP improves pose estimation, especially in challenging off-road scenarios where conventional rigid features are scarce. Furthermore, the analysis reveals that the effectiveness of this weighting strategy is highly environment-dependent, influenced by the structural and semantic composition of the scene.
\end{abstract}


\section{Introduction}

LiDAR-based odometry is a core component of autonomous robotic systems~\cite{jorge2024impact,barros2020improving}, enabling motion estimation in environments where reliable geometric correspondences are difficult to establish due to dynamic objects, sparse structure, vegetation, or unstructured terrain. By estimating the relative motion between consecutive point clouds, LiDAR odometry provides geometric constraints that are commonly integrated into localization, mapping, and SLAM frameworks for autonomous vehicles and mobile robots\cite{garrote2019mobile}.

Traditional LiDAR odometry methods rely mainly on geometric registration techniques, such as Iterative Closest Point (ICP)~\cite{besl1992method} and RANSAC~\cite{fischler1981random}. These methods estimate the rigid-body transformation between consecutive scans by minimizing residuals between corresponding points, features, or surface elements. Although such approaches have shown strong performance in structured urban environments, their accuracy often degrades in dynamic or unstructured scenarios.

A key limitation of geometry-only registration methods is the assumption that the environment is mostly static. In practice, many real-world environments contain dynamic or geometrically unstable elements, including moving vehicles, pedestrians, vegetation, tall grass, and deformable terrain. These elements can introduce incorrect correspondences during scan registration, leading to drift accumulation and degraded pose estimation.

\begin{figure}[t]
    \centering
    \includegraphics[width=1\columnwidth,trim={1.5cm 1cm 0cm 0cm},clip]{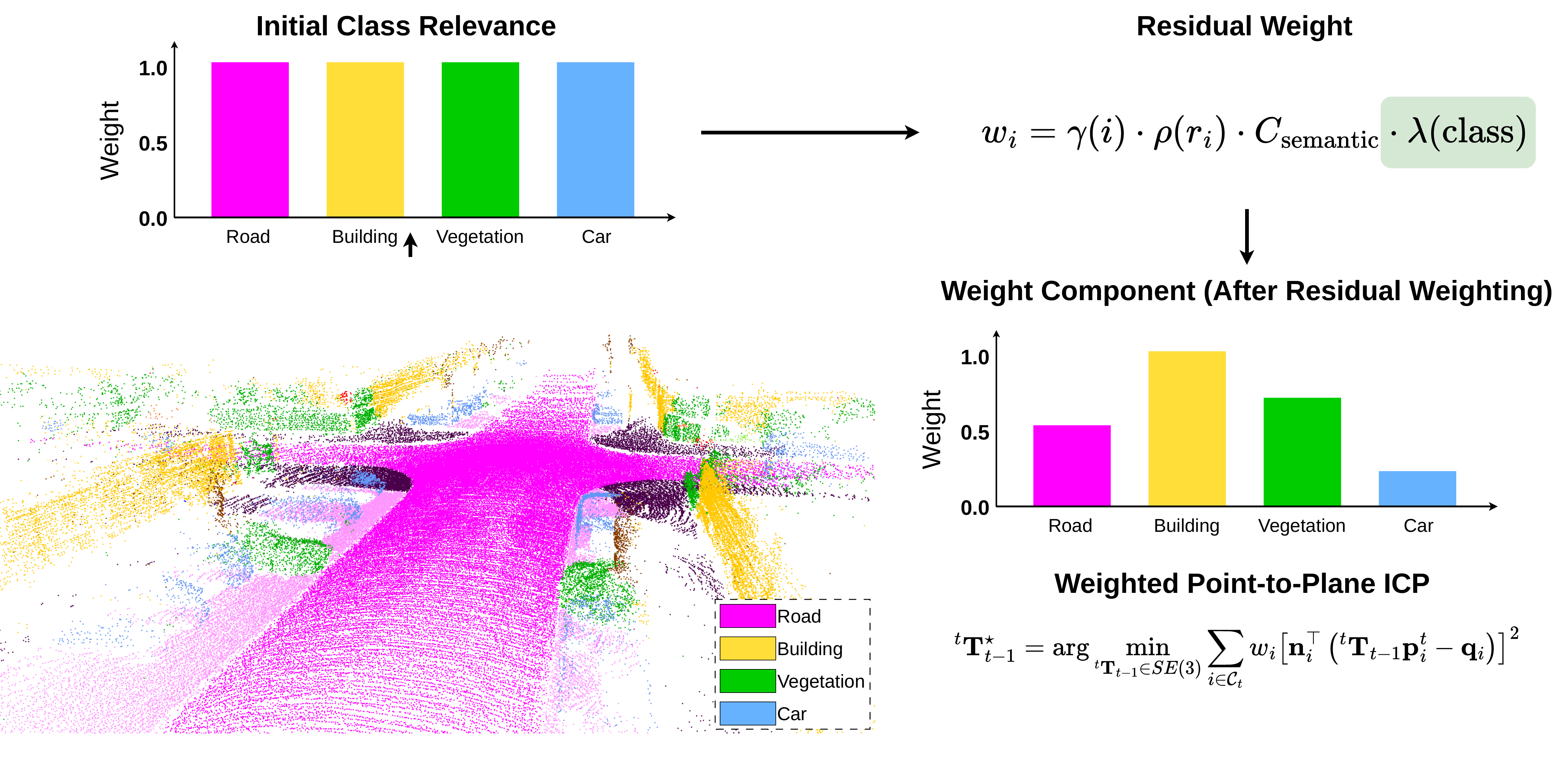}
    \caption{Conceptual illustration of the proposed semantic class-weighted ICP. Initially, all semantic classes contribute equally to the registration process. The proposed residual weighting assigns different class-dependent contributions before the weighted point-to-plane ICP optimization.}
    \label{fig:concept_weighting}
\end{figure}

To address this limitation, recent works have incorporated semantic information into the registration pipeline. Semantic cues can be used to identify scene elements and apply class-dependent operations at different stages of the odometry framework. Examples include class-based downsampling of the input scan~\cite{li2021saloam}, reducing the influence of dynamic objects in the map~\cite{chen2019sumapp}, and removing dynamic objects from the raw scan~\cite{du2021lidar}.

This work builds on semantic-based registration and argues that not all semantic classes contribute equally to accurate pose estimation. Static and geometrically stable structures, such as walls, are generally more reliable for scan registration than potentially dynamic or unstable elements, such as vehicles or vegetation. Based on this observation, we propose a semantic-weighted ICP formulation built on SuMa++~\cite{chen2019sumapp}. In addition to the Huber norm and semantic compatibility weights, the proposed method introduces a class-dependent weighting function that adjusts the contribution of each residual according to the semantic class of the associated points.

The proposed method is evaluated on two datasets: SemanticKITTI~\cite{behley2019iccv} and RELLIS-3D. SemanticKITTI~\cite{behley2019iccv} is a standard benchmark for odometry and SLAM in urban and suburban environments, whereas RELLIS-3D provides data from off-road environments. The results indicate that the proposed approach improves pose estimation compared with the SuMa++ baseline. The analysis also indicates that the effectiveness of semantic weighting depends on the environmental structure and the semantic distribution of the scene. In particular, off-road environments benefit from semantic weighting due to the prevalence of geometrically ambiguous terrain and vegetation.

The main contributions of this work are summarized as follows:

\begin{itemize}
    \item A semantic-weighted ICP approach that reduces the influence of potentially dynamic or geometrically unstable elements during scan registration;

    \item An experimental evaluation of the proposed Semantic-weighted ICP in urban and off-road environments;

    \item An analysis of the influence of different semantic classes on odometry performance across distinct environmental conditions.
\end{itemize}

\begin{figure*}[t]
    \centering
    \includegraphics[width=\textwidth,trim={0cm 0cm 4cm 5cm},clip]{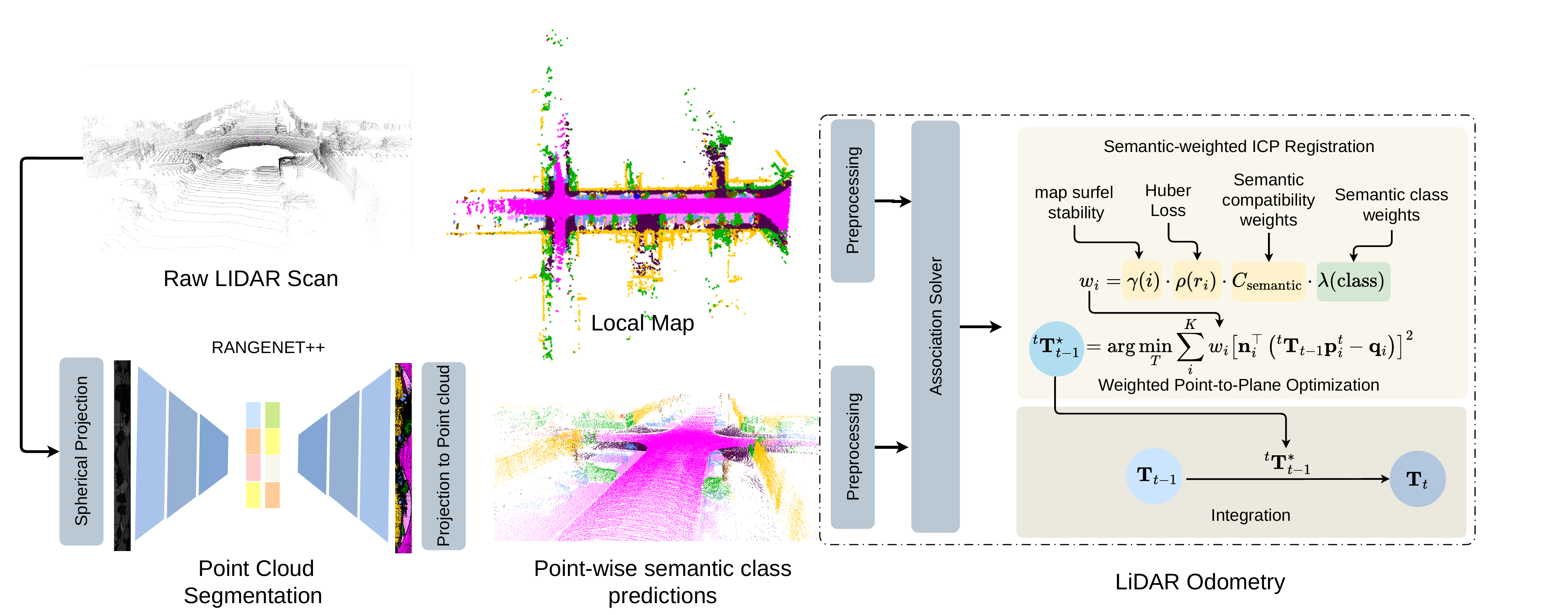} 
    \caption{Overview of the proposed LiDAR odometry pipeline. The raw LiDAR scan is processed by a semantic module (RangeNet++) to generate semantic labels. These labels, along with a class-aware weighting strategy, modulate the residual weights in a weighted point-to-plane ICP optimization before updating the surfel-based map.}
    \label{fig:pipeline_overview}
\end{figure*}

\section{Related Work}

\subsection{Geometric LiDAR Odometry}
LiDAR odometry is commonly formulated as a geometric registration problem, where motion is estimated by aligning consecutive scans or a scan to a local map~\cite{pomerleau2015review}. Classical ICP and its point-to-plane or generalized variants estimate rigid-body motion through iterative correspondence search and residual minimization~\cite{besl1992method,chen1992object,low2004linear,segall2009generalized}. Their performance, however, depends strongly on correspondence quality and the presence of stable geometric constraints~\cite{pomerleau2013comparing,pomerleau2015review}.

Feature-based methods, such as LOAM~\cite{zhang2014loam} and LeGO-LOAM~\cite{shan2018lego}, improve efficiency by registering selected edge, planar, or ground features. LiDAR-inertial methods, including LIO-SAM~\cite{shan2020lio} and FAST-LIO2~\cite{xu2022fast}, further improve robustness by coupling LiDAR with IMU measurements through factor-graph optimization or direct scan-to-map registration. Nevertheless, geometry-only front-ends remain sensitive to dynamic objects, vegetation, tall grass, irregular terrain, and other unstable structures that can introduce unreliable correspondences and increase drift~\cite{chen2019sumapp,pfreundschuh2021dynamic,kim2020remove,jiang2021rellis}.

\subsection{Semantic-aided LiDAR Odometry}
Semantic information has been incorporated into LiDAR odometry and SLAM to improve robustness in dynamic or weakly structured environments~\cite{chen2019sumapp,rosinol2020kimera,pan2021digging}. This has been enabled by annotated datasets such as SemanticKITTI~\cite{behley2019iccv}, nuScenes~\cite{caesar2020nuscenes}, and RELLIS-3D~\cite{jiang2021rellis}, as well as efficient LiDAR segmentation networks such as RangeNet++~\cite{milioto2019rangenet}, SalsaNext~\cite{cortinhal2020salsanext}, and Cylinder3D~\cite{zhu2021cylindrical}.

Existing methods use semantics mainly for point filtering, dynamic-object removal, semantic consistency in correspondence association, or semantic map representation~\cite{chen2019sumapp,dube2017segmatch,bowman2017probabilistic,pfreundschuh2021dynamic,kim2020remove,lim2021erasor}. For example, SuMa++~\cite{chen2019sumapp} integrates semantic segmentation into surfel-based mapping and projective ICP to reject dynamic objects and enforce label consistency. While effective, most approaches treat semantics primarily as a hard rejection or compatibility cue, rather than explicitly modeling the contribution of each class to the registration objective.

SuMa++~\cite{chen2019sumapp} extends surfel-based mapping by incorporating semantic segmentation into projective ICP. Semantic labels are used to reject dynamic objects and enforce semantic consistency during registration, improving robustness in dynamic urban environments. However, most semantic-aided methods primarily use semantics for hard rejection or label-consistent association, while the class-specific contribution to the registration objective is less explicitly modeled or evaluated~\cite{chen2019sumapp,pfreundschuh2021dynamic,lim2021erasor}.
This is limiting because the usefulness of a semantic class depends on both the environment and its geometric stability. Parked vehicles may provide useful constraints in urban scenes, whereas vegetation, tree trunks, grass, and terrain can dominate the available structure in off-road environments~\cite{jiang2021rellis,zhang2022semanticposs}. In contrast to hard filtering, this work studies semantic class-aware residual weighting in the ICP front-end, aiming to reduce the influence of unreliable structures while retaining classes that remain geometrically informative.

\section{Proposed Approach}

\subsection{Problem Formulation}
The aim of LiDAR odometry is to estimate a sequence of rigid-body transformations 
\begin{equation}
    \mathbf{T}_0, \mathbf{T}_1, \dots, \mathbf{T}_T, 
\end{equation}
\noindent where $\mathbf{T}_t = [\mathbf{R}_t,\mathbf{t}_t] \in SE(3)$ maps points from the LiDAR frame at time $t$ to a reference frame. 

In LiDAR odometry this mapping is computed based on consecutive LiDAR scans or between a LiDAR scan and a local map. For the purpose of this work we use local map $\mathcal{M}_t = \{\mathbf{q}_j  \in \mathbb{R}^3\}_{j=1}^{m_{t-1}}$ defined in a reference coordinate frame with points until $t$ .  Hence, let's define a sequence of point clouds 
\begin{equation}
\mathcal{P}_0, \mathcal{P}_1, \dots, \mathcal{P}_T, 
\end{equation}
\noindent where  each point cloud $\mathcal{P}_t = \{\mathbf{p}_i^t \in \mathbb{R}^3\}_{i=1}^{n_t}$ is defined in the LiDAR coordinate frame at time $t$. 
The LiDAR odometry problem is formulated as the estimation of ${}^{t}\mathbf{T}_{t-1} \in  SE(3)$ that best aligns the current scan $\mathcal{P}_t$  (transformed to the world coordinate frame using $\mathbf{T}_{t-1}$) and  the local map $\mathcal{M}_{t-1}$.

For the purpose of this work, and similar as in \cite{chen2019sumapp}, we use a point-to-plane registration algorithm:

\begin{equation}
{}^{t}\mathbf{T}_{t-1}^\star =
\argmin\limits_{{}^{t}\mathbf{T}_{t-1} \in SE(3)}
\sum_{i \in \mathcal{C}_t}
w_i \left[\mathbf{n}_i^\top \left( {}^{t}\mathbf{T}_{t-1} \mathbf{p}_i^t - \mathbf{q}_i \right) \right]^2),
\label{eq:point-to-plane}
\end{equation}
\noindent where
$\mathcal{C}_t$ denotes the set of valid point correspondences between $\mathcal{P}_t$ and $\mathcal{M}_{t-1}$ given by an association solver (see Fig.\ref{fig:pipeline_overview}), and $w_i$ corresponds to the residual weight.  For more details on point-to-plane registration, we refer the reader to \cite{pomerleau2015review}. 


Finally the global pose estimate $\mathbf{T}_t$ is updated recursively as follows:
\begin{equation}
\mathbf{T}_t =  \mathbf{T}_{t-1} ({}^{t}\mathbf{T}_{t-1}^\star)^{-1} \ ,
\qquad \mathbf{T}_0 = \mathbf{I}.
\end{equation}
\noindent where  $\mathbf{T}_{t-1}$ is the previous global transform and ${}^{t}\mathbf{T}_{t-1}^\star$ is obtained from \eqref{eq:point-to-plane}.

\subsection{Semantic-weighted ICP}
In this work we argue that not all elements in an environment are equally relevant for point cloud registration to achieve accurate alignment.  As such we propose a residual weighting function based on the semantic class of the corresponding points.  

Inspired by SuMa++~\cite{chen2019sumapp}, we adopt a similar framework in which RangeNet++~\cite{milioto2019rangenet} is used to generate semantic masks for both the incoming scan, $\mathcal{S}_P \in \mathbb{L}^{n_t}$, and the map, $\mathcal{S}_M \in \mathbb{L}^{m_{t-1}}$, where $\mathbb{L} = \{l_1,l_2,\dots,l_L\}$ denotes the set of $L$ semantic classes present in a given environment.

As for registration, we express \eqref{eq:point-to-plane} in vectorized form, yielding the following weighted error function:
\begin{equation}
E({}^{t}\mathbf{T}_{t-1})
=
\frac{1}{2}\mathbf{r}^\top
\mathbf{W}\mathbf{r},
\end{equation}
where 
$\mathbf{r} = [r_1,r_2,\dots,r_{|\mathcal{C}_t|}]^\top \in \mathbb{R}^{|\mathcal{C}_t|}$
is the stacked residual vector. Each residual is defined as
\begin{equation}
r_i = \mathbf{n}_i^\top
\left({}^{t}\mathbf{T}_{t-1}\mathbf{p}_i^t - \mathbf{q}_i \right),
\end{equation}
where $\mathbf{p}_i^t$ denotes a point in the current scan, $\mathbf{q}_i$ is its corresponding map point, and $\mathbf{n}_i$ is the associated normal. The matrix
$\mathbf{W} \in \mathbb{R}^{|\mathcal{C}_t| \times |\mathcal{C}_t|}$
is diagonal, with entries corresponding to the residual weights $w_i$. In this work, each weight $w_i$ is computed as
\begin{equation}
w_i = \hat{w}_i \cdot \lambda(\mathcal{S}_{P(i)}),
\end{equation}

where $\hat{w}_i$ follows the weighting formulation proposed in SuMa++~\cite{chen2019sumapp}, and is defined as follow:
\begin{equation}
\hat{w}_i =
\gamma(i)
\cdot \rho_{\mathrm{Huber}}(r_i)  \cdot
C_{\mathrm{sem}}(\mathcal{S}_{P(i)},\mathcal{S}_{M(i)}),
\end{equation}
\noindent  $\gamma(i)$ is an indicator function that returns $1$ if the corresponding map surfel is considered stable and $0$ otherwise,  $\rho_{\mathrm{Huber}}(\cdot)$ denotes the Huber loss function, and $C_{\mathrm{sem}}(\cdot,\cdot)$ is a semantic compatibility function. Specifically, $C_{\mathrm{sem}}$ returns the class likelihood $p_{c_i} \in [0,1]$ when the scan and map points belong to the same semantic class, and $1-p_{c_i}$ otherwise.

The term $\lambda(\mathcal{S}_{P(i)})$ denotes the proposed class-wise weighting function, which assigns an additional importance weight according to the semantic class of the source point:
\begin{equation}
\lambda:\mathbb{L}\longrightarrow \mathbb{R}^{+}.
\end{equation}
Thus, for each class $l_k \in \mathbb{L}$, the function assigns a positive scalar weight
\begin{equation}
\lambda(l_k) = w_{P_k}, \qquad w_{P_k} \in \mathbb{R}^{+},
\end{equation}
where $\mathbb{L}$ denotes the set of semantic classes. This formulation allows the registration objective to emphasize semantic classes that are expected to provide more reliable geometric constraints, while reducing the influence of classes that are less informative or more prone to dynamic behavior.

\begin{figure*}[ht]
\centering
\includegraphics[width=\textwidth]{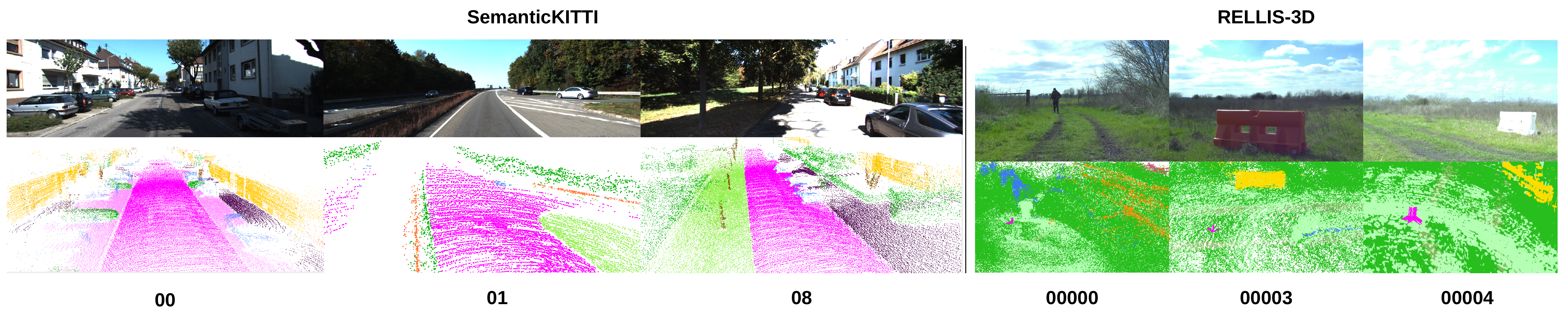}
\caption{Qualitative overview of the evaluation datasets. The top row displays representative RGB camera frames, while the bottom row illustrates the corresponding 3D point cloud semantic segmentation output. }
\label{fig:datasets_visual_overview}
\end{figure*}

\begin{table*}[ht]
\renewcommand{\arraystretch}{1.8}
\caption{Comparative Results on KITTI Odometry (Rotational Error ($^{\circ}/100$~m) / Translational Error \%)}
\label{tab:kitti_results}
\centering
\resizebox{\textwidth}{!}{%
\begin{tabular}{L{1cm} c c c c c c c c c c c c}
\tableline
Approach & 00* & 01 & 02* & 03 & 04 & 05* & 06* & 07* & 08* & 09* & 10 & Average  \\
& urban & h-way & urban & country & country & country & urban & urban & urban & urban & country & \\
\hline

SuMa++ & 0.23/0.66 & 0.47/2.01 & \textbf{0.38}/1.13 & \textbf{0.46/0.67} & 0.27/0.40 & 0.23/0.47 & \textbf{0.17/0.43} & 0.36/0.38 & \textbf{0.33/1.06} & 0.26/0.49 & 0.28/\textbf{0.65} & 0.310/0.759 \\ 
 & \multicolumn{12}{l}{All Semantic classes assigned equal reliability, no class-aware weighting applied.}\\ \hline

\multirow{3}{*}{\rotatebox[origin=r]{90}{
    \begin{tabular}{c}
    Semantic \\
    Weighted ICP
    \end{tabular}
    }}   
    & 0.23/0.66 & 0.94/22.69 & 0.83/2.19 & 0.65/0.79 & 0.30/0.85 & 0.18/0.38 & 0.27/0.49 & 0.36/0.47 & 0.35/1.07 & 0.20/0.62 & 0.31/0.72 & 0.42/2.81  \\ 
    & \multicolumn{12}{l}{Stable structures and roads high; ground/vegetation intermediate, parked vehicles/dynamic classes strongly suppressed.}  \\ 
    \cline{2-13}
    & 0.21/0.64 & 0.64/10.94 & 0.46/1.26 & 0.58/0.83 & \textbf{0.22}/0.58 & 0.18/0.38 & 0.28/0.50 & 0.41/0.45 & 0.38/1.15 & \textbf{0.15}/0.63 & 0.27/0.78 & 0.34/1.65 \\ 
    & \multicolumn{12}{l}{Stable structures further reinforced,road/ground/vegetation/cars intermediate, bikes and cyclists less suppressed.}  \\ 
    \cline{2-13}
    & \textbf{0.20/0.62} & \textbf{0.40/1.50} & \textbf{0.38/1.02} & 0.52/0.70 & \textbf{0.22/0.39} & \textbf{0.16/0.35} & 0.25/0.48 & 0.26/\textbf{0.36} & 0.34/1.08 & 0.16/\textbf{0.46} & 0.26/0.68 & \textbf{0.286/0.695} \\ 
    & \multicolumn{12}{l}{Stable structures high, road/ground/vegetation/cars intermediate, dynamic classes low.} \\ 
    \tableline
\end{tabular}
}
\begin{minipage}{\textwidth}
\vspace{5pt}
\footnotesize
Relative errors averaged over trajectories of 100 to 800 m length: relative rotational error in degrees per 100 m / relative translational error in \%. Sequences marked with an asterisk contain loop closures. Bold numbers indicate best performance in terms of translational error.
\end{minipage}
\end{table*}

\section{Experimental Evaluation}

The experimental evaluation was designed to assess the central hypothesis of this work: not all semantic entities in a scene contribute equally to scan registration. To this end, we evaluated the proposed approach on two datasets: SemanticKITTI~\cite{behley2019iccv} and RELLIS-3D~\cite{jiang2021rellis}. These datasets, illustrated in Fig.\ref{fig:datasets_visual_overview}, enable an analysis of Semantic-weighted ICP registration across diverse operating conditions, including urban, highway, rural, and off-road scenarios.

We use SuMa++~\cite{chen2019sumapp} as the baseline, where all semantic classes are treated with uniform importance during scan matching. To quantify the influence of class-dependent weighting on odometric accuracy, we evaluated several semantic weighting configurations. In these configurations, semantic classes are assigned low, intermediate, or high weights.

Odometry accuracy is evaluated using relative pose error (RPE), following a KITTI-style odometry evaluation protocol. We report relative translational drift as a percentage and relative rotational drift in degrees per $100$~m. These metrics characterize local trajectory consistency and enable comparison of drift across sequences and operating environments.

The experiments were conducted on a workstation equipped with an AMD Ryzen 7 3800X 8-Core CPU, 64~GB of RAM, and an NVIDIA GeForce RTX 3060 GPU with 12~GB of memory.

\subsection{Results and Discussion}
\begin{figure*}[t]
    \centering
    \includegraphics[width=0.95\textwidth, trim=0.5cm 0.5cm 2cm 0cm, clip]{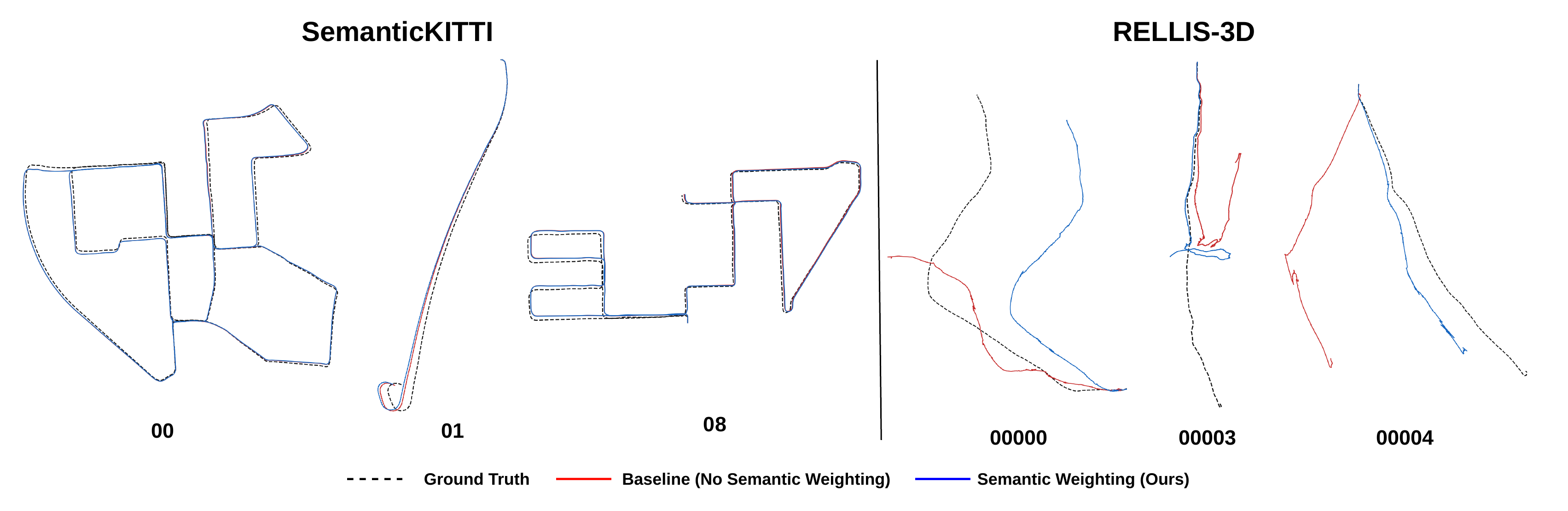}
    \caption{Qualitative trajectory comparison on representative SemanticKITTI~\cite{behley2019iccv} and RELLIS-3D sequences. The figure compares the ground truth trajectory, the baseline without semantic weighting, and the proposed semantic-weighted ICP.}
    \label{fig:trajectory_comparison}
\end{figure*}

In this section, we present and discuss  the performance of our semantic weighted ICP method across different environments. Table~\ref{tab:kitti_results} and  Table~\ref{tab:rellis_results} reports quantitative results  on SemanticKITTI~\cite{behley2019iccv} and  RELLIS-3D dataset, respectively, while Fig.\ref{fig:trajectory_comparison} presents qualitative results.

\begin{table*}[t]
\renewcommand{\arraystretch}{1.5}
\centering
\caption{Comparative Results on Rellis-3D Dataset (Rotational Error ($^{\circ}/100$~m) / Translational Error \%)}
\label{tab:rellis_results}
\resizebox{\textwidth}{!}{%
\begin{tabular}{l c c c c c c}
\tableline
Approach & 00000 & 00001 & 00002 & 00003 & 00004 & Average  \\
\hline
SuMa++ & 22.17 / 32.74 & 56.89 / 65.50 & 23.14 / 21.92 & 80.97 / 49.82 & 24.14 / 35.22 & 41.46 / 41.04 \\ 
&  \multicolumn{6}{l}{Semantic classes assigned equal reliability, no class-aware weighting applied}\\
\hline
\multirow{3}{*}{\rotatebox[origin=r]{90}{
    \begin{tabular}{c}
    Semantic \\
    Weighted ICP
    \end{tabular}
    }}   
& 12.54/25.39 & 98.80/79.91 & 26.22/29.19 & 67.07/45.28 & \textbf{7.26}/17.92 & 42.38/39.54 \\
&  \multicolumn{6}{l}{Stable structure further reinforced, natural elements and irregular terrain intermediate-low; vehicle low.} \\\cline{2-7}
& \textbf{7.84}/24.29 & 86.08/63.38 & 21.15/24.82 & \textbf{29.25}/34.96 & 15.17/33.11 & 31.90/36.11 \\
&  \multicolumn{6}{l}{Stable structures high, vegetation moderate, dynamic classes low.}\\ \cline{2-7}
& 10.76 / \textbf{17.20} & \textbf{45.32 / 49.78} & \textbf{16.88 / 21.26} & 46.02 /\textbf{ 33.07} & 9.77 / \textbf{11.80} & \textbf{24.67 / 25.77} \\ 
&\multicolumn{6}{l}{Balanced configuration, stable structures high, natural elements/terrain/roads intermediate; vehicle low.}\\
\tableline
\end{tabular}
}
\begin{minipage}{\textwidth}
\vspace{8pt}
\footnotesize
Relative errors for the Rellis-3D dataset: relative rotational error in degrees per 100 m / relative translational error in \%. Bold numbers indicate the best performance for both metrics in off-road environments.
\end{minipage}
\end{table*}

\subsubsection{Results on SemanticKITTI}

SemanticKITTI~\cite{behley2019iccv} contains urban, highway, and rural driving sequences with point-wise semantic annotations. It includes stable static structures, such as buildings, poles, traffic signs, and fences, as well as potentially dynamic objects, such as cars, trucks, buses, cyclists, and pedestrians.

The results indicate the proposed approach is sequence- and environment-dependent. The largest improvements occur in urban scenarios, where rigid and temporally stable structures provide reliable constraints for scan-to-map registration. In contrast, the effect is less consistent in highway and rural sequences, where the scene geometry is often more sparse, repetitive, or dominated by large planar surfaces.

Hard removal of potentially dynamic classes does not consistently improve odometry accuracy. Although filtering vehicles and other movable objects can reduce spurious correspondences, it may also discard useful geometric constraints, particularly in weakly structured scenes where these objects represent a significant part of the observable geometry. This can reduce correspondence density and degrade the conditioning of the ICP optimization.

The results therefore support a soft weighting strategy rather than hard semantic filtering. Potentially dynamic classes should be assigned low but non-zero weights, limiting their influence while preserving useful constraints from parked or slowly moving objects. Similarly, ground-related classes should be weighted carefully: they can introduce degeneracy when dominant, but also provide stable support when combined with vertical static structures.

\subsubsection{Results on RELLIS-3D}

RELLIS-3D contains off-road sequences in unstructured environments, including dirt roads, grass, vegetation, mud, trees, and sparse built structures. These conditions make scan-to-map registration challenging, since many correspondences arise from geometrically weak, non-rigid, or temporally unstable regions.

The results in Table~\ref{tab:rellis_results} show that semantic weighting has a stronger impact on RELLIS-3D than on SemanticKITTI~\cite{behley2019iccv}. In the baseline configuration, all points contribute equally to the ICP objective, regardless of semantic class. This leads to an average translational drift of $41.04\%$ and a rotational drift of $41.46^{\circ}/100$~m across the five sequences, indicating that uniform weighting is poorly suited to off-road environments.

Reducing the influence of vegetation- and terrain-related classes improves registration stability. However, these classes cannot be removed entirely, since they account for a large fraction of the observable geometry. Unlike urban scenes, where buildings, poles, traffic signs, and road boundaries provide strong geometric constraints, off-road scenes often contain few rigid structures. Excessively down-weighting natural classes can therefore reduce correspondence density and make the ICP optimization under-constrained.

The best results are obtained with weights that balance class relevance. Sparse but rigid classes, such as buildings, poles, barriers, and logs, are assigned higher weights due to their geometric repeatability. Natural classes, such as grass, trees, bushes, and dirt, are assigned intermediate or low weights, allowing them to contribute while limiting the effect of noisy correspondences. Unreliable or non-informative classes, such as vehicles, people, void, and sky, are strongly attenuated or ignored.

\section{Conclusion}

This work explored semantic weighted ICP for LiDAR odometry across environments. The proposed approach extends the ICP registration stage of SuMa++ by assigning class-dependent relevance weights to semantic categories, allowing each class to contribute differently to scan-to-map alignment.

The results on SemanticKITTI~\cite{behley2019iccv} and RELLIS-3D show that semantic weighting can improve odometry accuracy, but its effectiveness depends on the operating environment and on the geometric role of each semantic class. In structured urban scenes, stable and geometrically salient classes, such as buildings, poles, and road-related structures, provide reliable constraints. In off-road scenes, natural classes, such as grass, trees, and terrain, require more careful weighting because they may introduce noisy correspondences while still representing a large fraction of the observable geometry.

Overall, the results indicate that semantic information should be used as a class-dependent reliability prior in the ICP objective, rather than as a binary filtering mechanism. The best configurations assign high weights to stable and geometrically salient classes, intermediate weights to ground-, vegetation-, and terrain-related classes, and low but non-zero weights to potentially dynamic objects. This strategy preserves useful geometric constraints while reducing the influence of unstable correspondences.

\section*{Acknowledgment}
This work has been supported by the project PharmaRobot (ref. COMPETE2030-FEDER-01478600), and funded by Fundação para a Ciência e a Tecnologia (FCT), Portugal. This work was also supported by the ISR-UC FCT grant UID/00048/2025 (DOI: 10.54499/UIDB/00048/2025)

\bibliographystyle{IEEEtran}
\bibliography{ref,own}

\end{document}